\documentclass[11pt]{article}

\usepackage[final]{acl}

\usepackage{times}
\usepackage{latexsym}
\usepackage{booktabs}
\usepackage{amssymb}
\usepackage{booktabs}
\usepackage{multirow}
\usepackage{adjustbox}
\usepackage{tabularx}
\usepackage{enumitem}
\usepackage{tcolorbox}
\tcbuselibrary{skins,breakable}
\usepackage[table,xcdraw]{xcolor}
\usepackage{fix-cm}
\usepackage{soul}
\usepackage[T1]{fontenc}

\usepackage[utf8]{inputenc}

\usepackage{microtype}

\usepackage{inconsolata}

\usepackage{graphicx}

\usepackage{xspace}
\newcommand{\dataset}{ToolHACE\xspace}

\usepackage{tikz}
\usepackage{caption}

\definecolor{CardBg}{HTML}{F3F4F6}
\definecolor{InnerBg}{HTML}{FFFFFF}
\definecolor{LineGray}{HTML}{D6D9DE}
\definecolor{TextGray}{HTML}{3F3F46}
\definecolor{TitleGray}{HTML}{2F3437}

\definecolor{BlueA}{HTML}{2563EB}
\definecolor{OrangeA}{HTML}{D97706}
\definecolor{RedA}{HTML}{BE123C}
\definecolor{GreenA}{HTML}{15803D}
\definecolor{PurpleA}{HTML}{7C3AED}
\definecolor{TealA}{HTML}{0F766E}

\newcommand{\emoji}[1]{\includegraphics[height=1em]{#1}}

\tcbset{
  cardbox/.style={
    enhanced,
    arc=1.6mm,
    boxrule=0pt,
    colback=CardBg,
    left=4pt,right=4pt,top=4pt,bottom=4pt,
    width=\linewidth,
    height=6.3cm,
    valign=top,
  },
  routebox/.style 2 args={
    enhanced,
    arc=1mm,
    boxrule=.3pt,
    colback=#1!7,
    colframe=#1!55!black,
    left=3pt,right=3pt,top=2pt,bottom=2pt,
    title={#2},
    fonttitle=\bfseries\fontsize{5.8}{6.2}\selectfont,
    coltitle=#1!85!black,
    colbacktitle=#1!12,
  }
}

\newcommand{\smalltxt}[1]{{\fontsize{6.4}{6.9}\selectfont\textcolor{TextGray}{#1}}}
\newcommand{\tinytxt}[1]{{\fontsize{5.4}{5.8}\selectfont\textcolor{TextGray}{#1}}}

\newcommand{\iconbadge}[2]{%
\tikz[baseline=-0.55ex]\node[
  circle,
  fill=#1!14,
  draw=#1!70!black,
  line width=.28pt,
  minimum size=1.75em,
  inner sep=0pt,
  font=\fontsize{5.5}{5.8}\selectfont\bfseries,
  text=#1!80!black
]{#2};%
}

\newcommand{\stepheader}[3]{%
\noindent\iconbadge{#1}{#2}\hspace{0.25em}%
{\bfseries\fontsize{6.6}{7.0}\selectfont\textcolor{TitleGray}{#3}}%
}

\newcommand{\pill}[2]{%
\tcbox[
  on line,
  arc=.8mm,
  boxrule=.25pt,
  colback=#1!10,
  colframe=#1!50!black,
  left=1.8pt,right=1.8pt,top=.6pt,bottom=.6pt
]{\fontsize{5.4}{5.8}\selectfont\textcolor{#1!80!black}{\textbf{#2}}}%
}

\tcbset{
  flowbox/.style={
    enhanced,
    arc=1mm,
    boxrule=.3pt,
    colback=BlueA!8,
    colframe=BlueA!50!black,
    left=2pt,
    right=2pt,
    top=1.5pt,
    bottom=1.5pt,
    width=\linewidth,
  }
}

\title{ToolHACE: A Dataset for Detecting Hallucinations in \\ Tool-Based LLM Responses}

\author{First Author \\
  Affiliation / Address line 1 \\
  Affiliation / Address line 2 \\
  Affiliation / Address line 3 \\
  \texttt{email@domain} \\\And
  Second Author \\
  Affiliation / Address line 1 \\
  Affiliation / Address line 2 \\
  Affiliation / Address line 3 \\
  \texttt{email@domain} \\}

\author{
 \textbf{Nikolay Ivanov\textsuperscript{1,*}},
 \textbf{Sergey Pletenev\textsuperscript{1,2,*}},
 \textbf{Elisei Rykov\textsuperscript{1,4*}},
 \textbf{Anatolii Frolov\textsuperscript{1,2,*}},
\\
 \textbf{Bogdan Monogov\textsuperscript{1}},
 \textbf{Alexander Potekhin\textsuperscript{1}},
 \textbf{Maria Marina\textsuperscript{1,2}},
 \textbf{Daniil Moskovskiy\textsuperscript{1,2}},
\\
 \textbf{Viktor Moskvoretskii\textsuperscript{3}},
 \textbf{Alexander Panchenko\textsuperscript{1,2}}, and
 \textbf{Ilseyar Alimova\textsuperscript{1}}
\\
 \textsuperscript{1}Skoltech,
 \textsuperscript{2}AIRI,
 \textsuperscript{3}EPFL,
 \textsuperscript{4}T-Tech
\\
 \small{
   \textbf{Correspondence:} \href{mailto:a.panchekno@skoltech.ru}{\{nikolay.ivanov, a.panchenko, i.alimova\}@skol.tech}
 }
}
\begin{document}
\maketitle
\begin{abstract}
\begingroup
\renewcommand\thefootnote{*}
\footnotetext{These authors contributed equally to this work.}
\endgroup
Tool-augmented language models ground their responses in external tool outputs, improving reliability beyond parametric knowledge alone. However, even when the correct tool is called and returns accurate information, models may still contradict such outputs, omit critical details, or generate unsupported claims. Existing hallucination benchmarks for tool-augmented dialogue primarily focus on the tool-calling stage, evaluating whether the correct tool and arguments are selected, while overlooking errors that arise during post-tool response generation. To address this gap, we introduce \dataset, a span-level benchmark for hallucination detection in tool-augmented LLM responses, covering five fine-grained hallucination categories. Through extensive experiments with modern hallucination detectors, we show that post-tool-call hallucinations are widespread and remain challenging to detect. Our results further demonstrate that effective detection requires training on specialized tool-augmented corpora, while transfer from tool-agnostic hallucination datasets performs poorly in this setting. These findings highlight the need for dedicated benchmarks and models for reliable hallucination detection in tool-augmented language systems.
\end{abstract}

\section{Introduction}


\begin{figure}[t]
    \centering
    \includegraphics[width=\columnwidth]{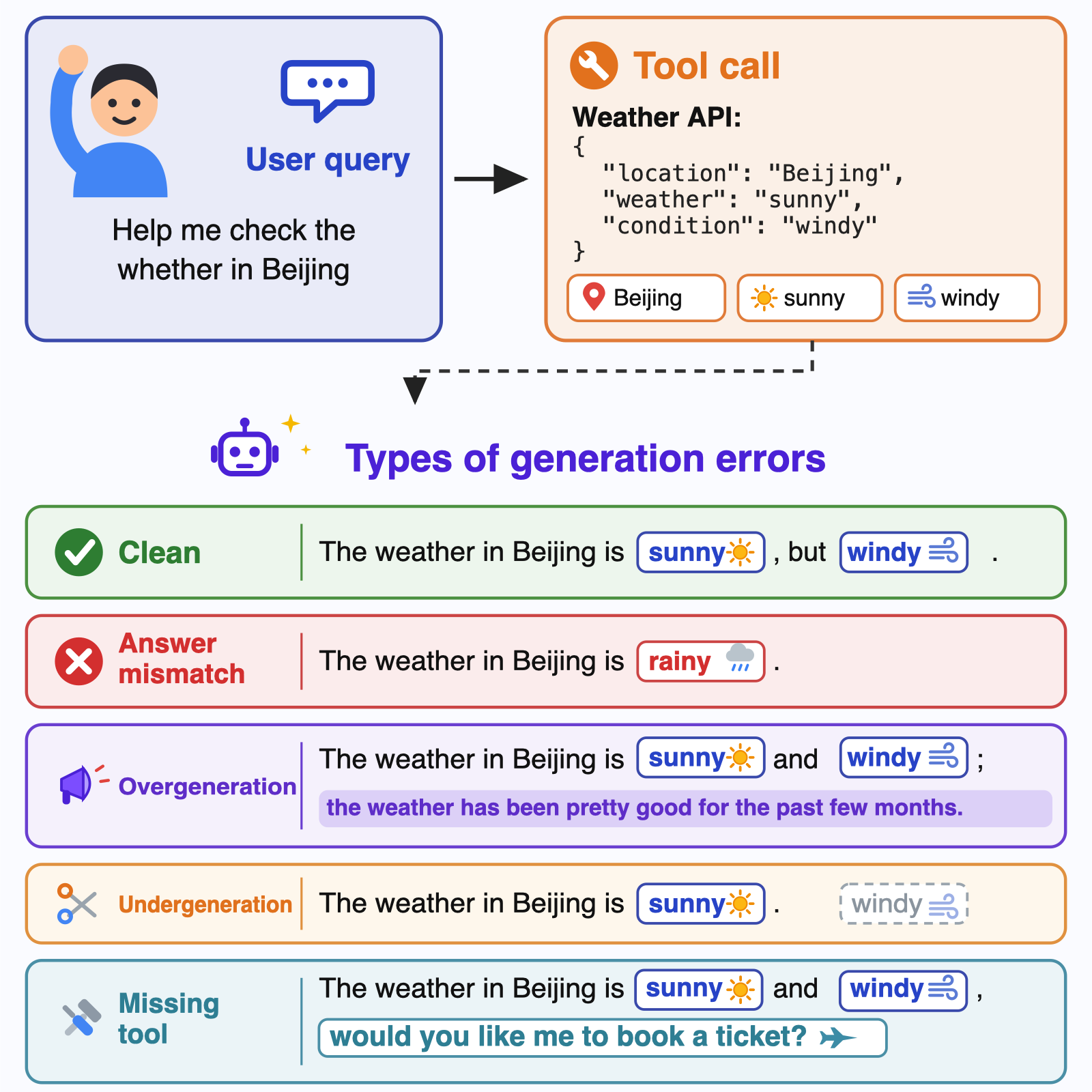}
    \caption{Main types of generation errors in tool-augmented LLM responses. A model can produce a correct answer, contradict the tool output, add unsupported information, omit a tool-provided fact, or introduce an unrelated tool/action request.}
    \label{fig:intro}
\end{figure}

Large language models (LLMs) are increasingly used as the reasoning core of tool-augmented agents, where the model decides when to call external APIs, search engines, calculators, or databases, and then synthesizes a final natural-language response based on the tool outputs \citep{patil2024gorilla, patil2025bfcl}. During the operation of such systems, various errors may arise, including incorrect tool selection (e.g., calling a search engine instead of a calculator), wrong tool arguments (malformed or incomplete inputs), missing tool invocation when external tools are required, incorrect tool-use sequencing leading to inconsistent or unusable outputs, and incorrect processing of tool outputs, where the model misinterprets or fails to properly incorporate the tool response. 

Most existing work on tool-augmented LLMs focuses on building large-scale benchmarks for tool calling itself, evaluating whether the model selects the correct API and provides valid arguments, while treating the final natural-language response as out of scope~\citep{patil2024gorilla,liu2024toolacewinningpointsllm,patil2025bfcl,ye2025toolhop,huang2024metatool,zhang2024toolbehonest}. However models may still hallucinate in their final response and fail to properly utilize the tool output, resembling retrieval augmented generation (RAG) pitfalls. We refer to these as \emph{post-tool-call hallucinations}: errors that occur \emph{after} the right tool has been invoked, in the natural-language response that the user actually reads. Detecting them requires evaluating the final response against the returned tool output, which is not addressed by existing tool-calling benchmarks.

To evaluate how well models can detect such hallucinations, we introduce \dataset, a dataset for detecting ungrounded generations in tool-augmented LLMs. Our dataset is built on top of ToolACE~\citep{liu2024toolacewinningpointsllm} and annotated at the span level with five categories: \emph{Correct}, \emph{Answer Mismatch}, \emph{Overgeneration}, \emph{Undergeneration}, and \emph{Missing Tool}, as illustrated in Figure~\ref{fig:intro}. We benchmark a range of modern hallucination detectors, including zero-shot prompted LLMs, encoder-based models trained on general-purpose hallucination corpora, and detectors fine-tuned on our dataset. Our results show that (i) post-tool-call hallucinations are widespread in modern tool-augmented LLMs, (ii) detectors trained solely on tool-agnostic hallucination data transfer poorly to this setting, and (iii) training on in-domain tool-calling data substantially reduces this performance gap.

Our contributions are:
\begin{itemize}[noitemsep,topsep=2pt]
\item We introduce a new span-level dataset for detecting hallucinations in the post-tool-call stage of tool-augmented LLMs.
\item We benchmark a range of state-of-the-art methods and modern LLM-based hallucination detectors on the proposed dataset.
\item We evaluate the effectiveness of transferring knowledge from hallucination detection in RAG systems to the tool-augmented setting.
\end{itemize}

\begin{figure*}[t]
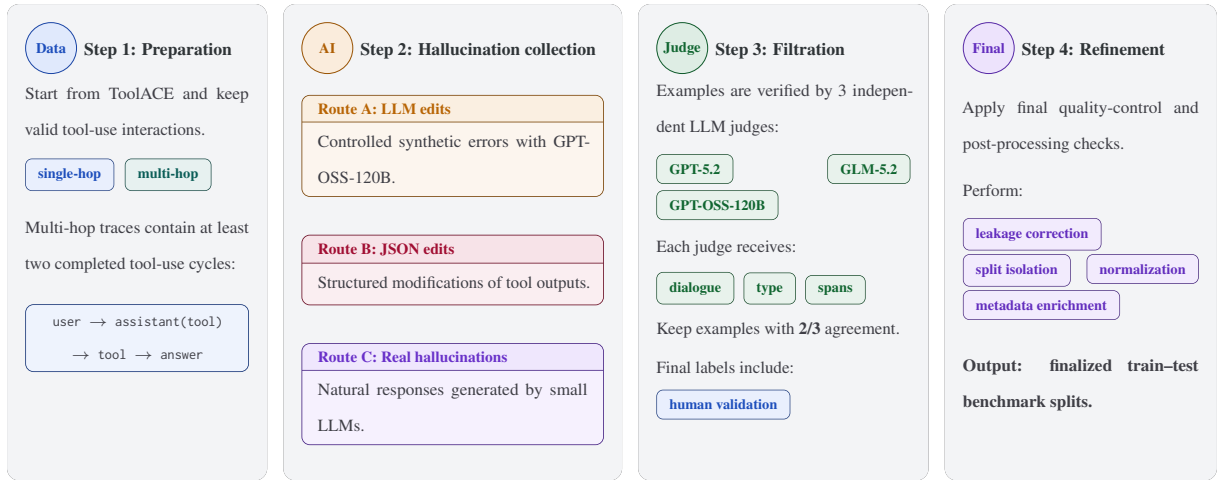

\centering

\begin{minipage}[t]{0.215\textwidth}
\begin{tcolorbox}[cardbox]
\stepheader{BlueA}{Data}{Step 1: Preparation}

\vspace{0.25em}
\smalltxt{Start from ToolACE and keep valid tool-use interactions.}

\vspace{0.25em}
\smalltxt{
\pill{BlueA}{single-hop}
\hspace{0.12em}
\pill{TealA}{multi-hop}
}

\vspace{0.25cm}
\smalltxt{Multi-hop traces contain at least two completed tool-use cycles:}

\vspace{0.2em}

\begin{tcolorbox}[flowbox]
\centering
\tinytxt{\texttt{user $\rightarrow$ assistant(tool)}}\\[-1pt]
\tinytxt{\texttt{$\rightarrow$ tool $\rightarrow$ answer}}
\end{tcolorbox}

\end{tcolorbox}
\end{minipage}%
\hfill%
\begin{minipage}[t]{0.28\textwidth}
\begin{tcolorbox}[cardbox]
\stepheader{OrangeA}{AI}{Step 2: Hallucination collection}

\vspace{0.10em}

\begin{tcolorbox}[routebox={OrangeA}{Route A: LLM edits}]
\smalltxt{Controlled synthetic errors with GPT-OSS-120B.}

\end{tcolorbox}

\vspace{0.06em}

\begin{tcolorbox}[routebox={RedA}{Route B: JSON edits}]
\smalltxt{Structured modifications of tool outputs.}

\end{tcolorbox}

\vspace{0.06em}

\begin{tcolorbox}[routebox={PurpleA}{Route C: Real hallucinations}]
\smalltxt{Natural responses generated by small LLMs.}

\end{tcolorbox}

\end{tcolorbox}
\end{minipage}%
\hfill%
\begin{minipage}[t]{0.24\textwidth}
\begin{tcolorbox}[cardbox]
\stepheader{GreenA}{Judge}{Step 3: Filtration}

\vspace{0.15em}

\smalltxt{
Examples are verified by 3 independent LLM judges:
}


\vspace{0.2em}
\smalltxt{
\pill{GreenA}{GPT-5.2}
\hspace{0.02em}
\pill{GreenA}{GLM-5.2}
\hspace{0.02em}
\pill{GreenA}{GPT-OSS-120B}
}

\vspace{0.2em}
\smalltxt{Each judge receives:}

\vspace{0.12em}
\smalltxt{
\pill{GreenA}{dialogue}
\hspace{0.05em}
\pill{GreenA}{type}
\hspace{0.05em}
\pill{GreenA}{spans}
}

\vspace{0.2em}
\smalltxt{Keep examples with \textbf{2/3} agreement.}

\vspace{0.12em}
\smalltxt{Final labels include:}

\pill{BlueA}{human validation}

\end{tcolorbox}
\end{minipage}%
\hfill%
\begin{minipage}[t]{0.225\textwidth}
\begin{tcolorbox}[cardbox]
\stepheader{PurpleA}{Final}{Step 4: Refinement}

\vspace{0.75em}
\smalltxt{Apply final quality-control and post-processing checks.}

\vspace{0.35em}
\smalltxt{Perform:}

\vspace{0.2em}
\smalltxt{
\pill{PurpleA}{leakage correction}
\hspace{0.1em}
\pill{PurpleA}{split isolation}
\hspace{0.1em}
\pill{PurpleA}{normalization}
\hspace{0.1em}
\pill{PurpleA}{metadata enrichment}
}

\vspace{0.85em}
\smalltxt{\textbf{Output: finalized train--test benchmark splits.}}

\end{tcolorbox}
\end{minipage}

\caption{
Dataset generation pipeline. ToolACE dialogues are split into single-hop and multi-hop traces.  Hallucinations are collected through LLM-based
synthetic edits, programmatic JSON perturbations, and naturally occurring
(\emph{wild}) generations from small LLMs. The resulting examples are
filtered by multiple judges and subsequently refined through quality-control,
split isolation, and normalization.
}
\label{fig:dataset_pipeline_compact}
\end{figure*}

\section{Related Work}

\begin{table}[!t]
\centering
\small
\resizebox{1\columnwidth}{!}{
\begin{tabular}{lcc}
\toprule
\textbf{Resource} & \textbf{Tool-calling} & \textbf{Span-level} \\
                  & \textbf{setting}      & \textbf{hallucination labels} \\
\midrule
ToolBench, BFCL, ToolACE, APIBench,    & \emoji{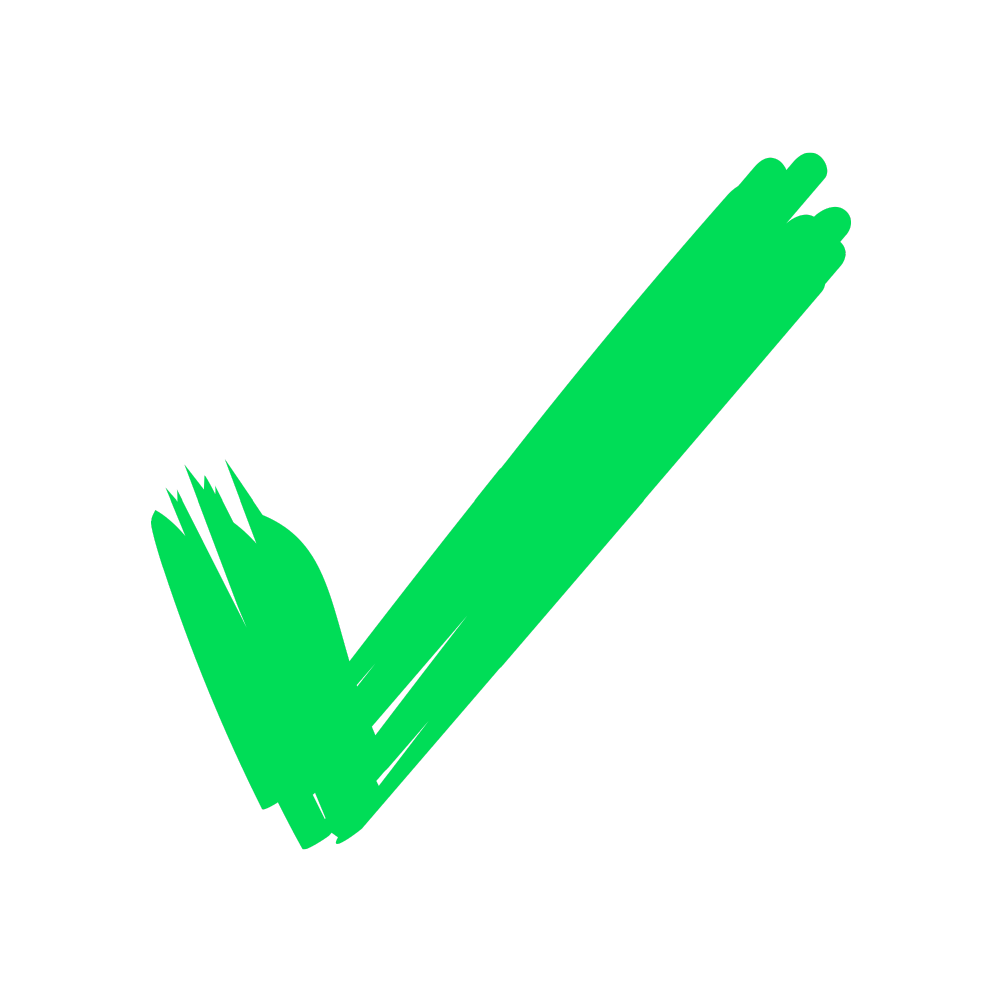} & \emoji{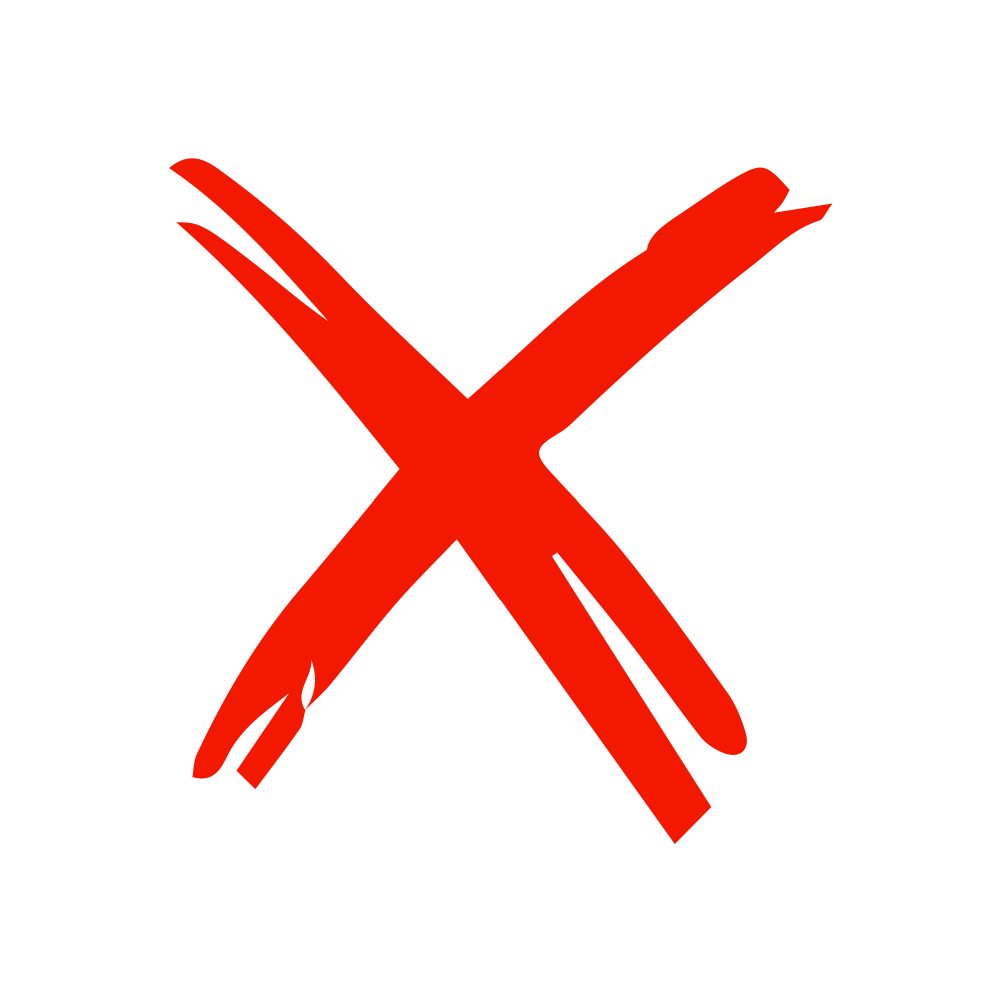} \\
APIGen, ToolAlpaca, StableToolBench, &   \emoji{images/green_ok.pdf} & \emoji{images/red_cross.pdf} \\
ToolHop, T-Eval, MetaTool              & \emoji{images/green_ok.pdf} & \emoji{images/red_cross.pdf} \\
ToolEmu, InjecAgent, ToolBH         & \emoji{images/green_ok.pdf} & \emoji{images/red_cross.pdf} \\
\midrule
RAGTruth, FAVA, PsiloQA, HAD                         & \emoji{images/red_cross.pdf}        & \emoji{images/green_ok.pdf} \\
\midrule
\textbf{Ours (\dataset)}                          & \emoji{images/green_ok.pdf} & \emoji{images/green_ok.pdf} \\
\bottomrule
\end{tabular}
}
\caption{Existing resources cover either the tool-calling setting or span-level hallucination annotation, but not both. Our resource combines them.}
\label{tab:related}
\end{table}

\paragraph{Tool-calling datasets.}
Tool-use benchmarks evaluate whether models select the correct tool and pass valid arguments, scored by AST match, executable correctness, or task pass rate~\cite{patil2024gorilla, patil2025bfcl, liu2024toolacewinningpointsllm, ye2025toolhop}. Scalable synthetic pipelines have expanded coverage to tens of thousands of dialogues~\cite{tang2023toolalpaca, liu2024apigen}. A smaller line of work expands the threat model: MetaTool~\cite{huang2024metatool} tests abstention when no appropriate tool exists; ToolEmu~\cite{ruan2024identifying} simulates execution for safety; InjecAgent~\cite{zhan2024injecagent} measures prompt-injection susceptibility; and ToolBH~\cite{zhang2024toolbehonest} annotates tool-calling hallucination at the tool-name level. Across this family, evaluation centers on the tool call itself; the natural-language response synthesized after tool execution is treated as out of scope, and span-level hallucination annotations are absent throughout.

\paragraph{Span-level hallucination detection.}
A separate line of work provides fine-grained span- or word-level hallucination annotations, but in non-tool-calling settings. RAGTruth~\cite{ragtruth} annotates RAG outputs across QA, summarization, and data-to-text (with tabular rather than tool-call inputs); FAVA~\cite{fava} synthesizes spans from a fine-grained taxonomy; PsiloQA~\cite{psiloqa} extends span annotation multilingually; and HAD~\cite{xu2025hadhallucinationdetectionlanguagse} offers an 11-category hierarchical taxonomy over twelve NLG tasks. None of these contain structured tool outputs as input, and none include an analogue of our \emph{Missing Tool} class. Models for this task fall into two camps: generative detectors that emit span markup~\cite{ragtruth, fava}, and encoder-based token classifiers~\cite{Kovacs:2025, modernbert-32k-haldetect}; we evaluate representatives of both. Table~\ref{tab:related} summarizes how existing resources cover one axis or the other but not both.

\section{The \dataset Dataset}

\subsection{Hallucination Types}

The dataset contains five hallucination labels, including the grounded class \emph{Correct} and four post-tool-call hallucination categories: \emph{Answer Mismatch}, \emph{Overgeneration}, \emph{Undergeneration}, and \emph{Missing Tool}. \emph{Undergeneration} is not annotated at the span level, whereas the other three hallucination categories have corresponding span-level annotations. Detailed definitions of each hallucination category are presented in Table~\ref{tab:classes}. Three of these (Answer Mismatch, Overgeneration, Undergeneration) have rough analogues in prior hallucination taxonomies such as HAD~\cite{xu2025hadhallucinationdetectionlanguagse} and RAGTruth~\cite{ragtruth}. \emph{Missing Tool}, however, is specific to tool-augmented settings and, to the best of our knowledge, has no direct counterpart in existing benchmarks. It captures cases where a model fabricates tool-mediated capabilities, such as claiming to have booked a reservation, retrieved a live price, or sent a message when no such tool exists or was called. We describe the synthesis pipeline for each non-Correct class below. 

\begin{table}[!t]
    \centering
    \footnotesize
    \begin{tabularx}{\linewidth}{@{} l >{\raggedright\arraybackslash}X @{}}
    \toprule
        Class & Definition \\ \midrule
        Correct & Answer is faithful to the tool response and fully addresses the query. \\ \midrule
        Answer Mismatch & Answer asserts a value or fact that the tool response \emph{contradicts}. \\ \midrule
        Overgeneration & Answer adds content with no support in the tool response (e.g.\ elaboration, background, general knowledge), without claiming any tool-mediated action. \\ \midrule
        Missing Tool & Answer references an \emph{action or capability} that no available tool can supply (e.g.\ offering to book, send, fetch, or compute when no such tool exists). \\ \midrule
        Undergeneration & Answer omits information from the tool response that the query required. \\ \bottomrule
    \end{tabularx}
    \caption{The five classes in \dataset. Boundaries are designed to be mutually exclusive: \emph{Answer Mismatch} requires a contradicting value in the tool response; \emph{Overgeneration} adds free-form content; \emph{Missing Tool} specifically claims tool-mediated capability.}
    \label{tab:classes}
\end{table}

\subsection{Source Corpus and Data Preparation}

We build \dataset on ToolACE, a large-scale tool-calling corpus spanning diverse APIs and application domains~\cite{liu2024toolacewinningpointsllm}. Compared to existing tool-calling benchmarks, ToolACE is particularly well suited for our setting due to its structured interaction traces comprising user queries, assistant tool invocations, and corresponding tool outputs. This explicit grounding structure enables fine-grained analysis of model behavior after tool execution. In addition, its broad coverage of heterogeneous tools and realistic usage scenarios increases the diversity of the resulting benchmark.

Each ToolACE sample serves as a seed instance for our construction pipeline. The original grounded answer is retained as a positive \emph{Correct} example, while the associated question--tool-response pair is used to synthesize one or more hallucinated variants.

We start from the full ToolACE corpus of 11,300 dialogues and retain only samples containing at least one complete tool-use interaction of the form: user query $\rightarrow$ assistant tool call $\rightarrow$ tool response $\rightarrow$ assistant answer. The retained dialogues are then divided into two subsets. \emph{Singlehop} dialogues contain exactly one tool invocation, whereas \emph{multihop} dialogues contain at least two complete tool-use cycles.

For multihop dialogues, we remove trailing follow-up questions to ensure that the final assistant response is self-contained. For singlehop dialogues, we observe that many samples terminate immediately after the tool call without an explicit tool response or final assistant answer. For these cases, we synthetically generate the missing tool response and grounded completion using an LLM, followed by structural filtering to remove malformed examples and multihop artifacts. The resulting dataset therefore combines real and synthetically completed singlehop and multihop instances while preserving traceability to the original ToolACE source through \texttt{dialogue\_id} and \texttt{source\_row\_index}.

\subsection{Hallucination Insertion Pipeline}
\label{subsec:data-generation}

\definecolor{TypeIcol}{RGB}{196,30,58}
\definecolor{TypeIbg}{RGB}{255,232,232}
\definecolor{TypeIIcol}{RGB}{21,101,192}
\definecolor{TypeIIbg}{RGB}{219,234,254}
\definecolor{TypeIIIcol}{RGB}{109,40,217}
\definecolor{TypeIIIbg}{RGB}{237,233,254}
\definecolor{jsonbg}{RGB}{243,244,246}
\definecolor{querybg}{RGB}{249,250,251}
\definecolor{darkgray}{RGB}{75,85,99}
\definecolor{labelgray}{RGB}{107,114,128}

\newcommand{\highlight}[1]{%
  \sethlcolor{TypeIbg}%
  \hl{\textcolor{TypeIcol}{#1}}%
}

\newcommand{\hspan}[3]{%
  \colorbox{#1}{\textcolor{#2}{\strut\textbf{#3}}}%
}
\newcommand{\spanI}[1]{\hspan{TypeIbg}{TypeIcol}{#1}}
\newcommand{\spanII}[1]{\hspan{TypeIIbg}{TypeIIcol}{#1}}
\newcommand{\spanIII}[1]{\hspan{TypeIIIbg}{TypeIIIcol}{#1}}
\tcbset{
  hallex/.style 2 args={
    enhanced, breakable, sharp corners=south,
    colback=white, colframe=#1, boxrule=1.2pt,
    attach boxed title to top left={xshift=8pt, yshift=-\tcboxedtitleheight/2},
    boxed title style={
      colback=#1, colframe=#1, arc=3pt,
      left=6pt, right=6pt, top=2pt, bottom=2pt,
      fontupper=\small\bfseries\color{white},
    },
    title={#2},
    top=10pt, bottom=6pt, left=8pt, right=8pt,
  },
  jsonbox/.style={
    colback=jsonbg, colframe=jsonbg, arc=3pt,
    left=6pt, right=6pt, top=3pt, bottom=3pt,
    fontupper=\footnotesize\ttfamily,
    top=3pt, bottom=3pt,
  },
}
\begin{figure*}[ht!]
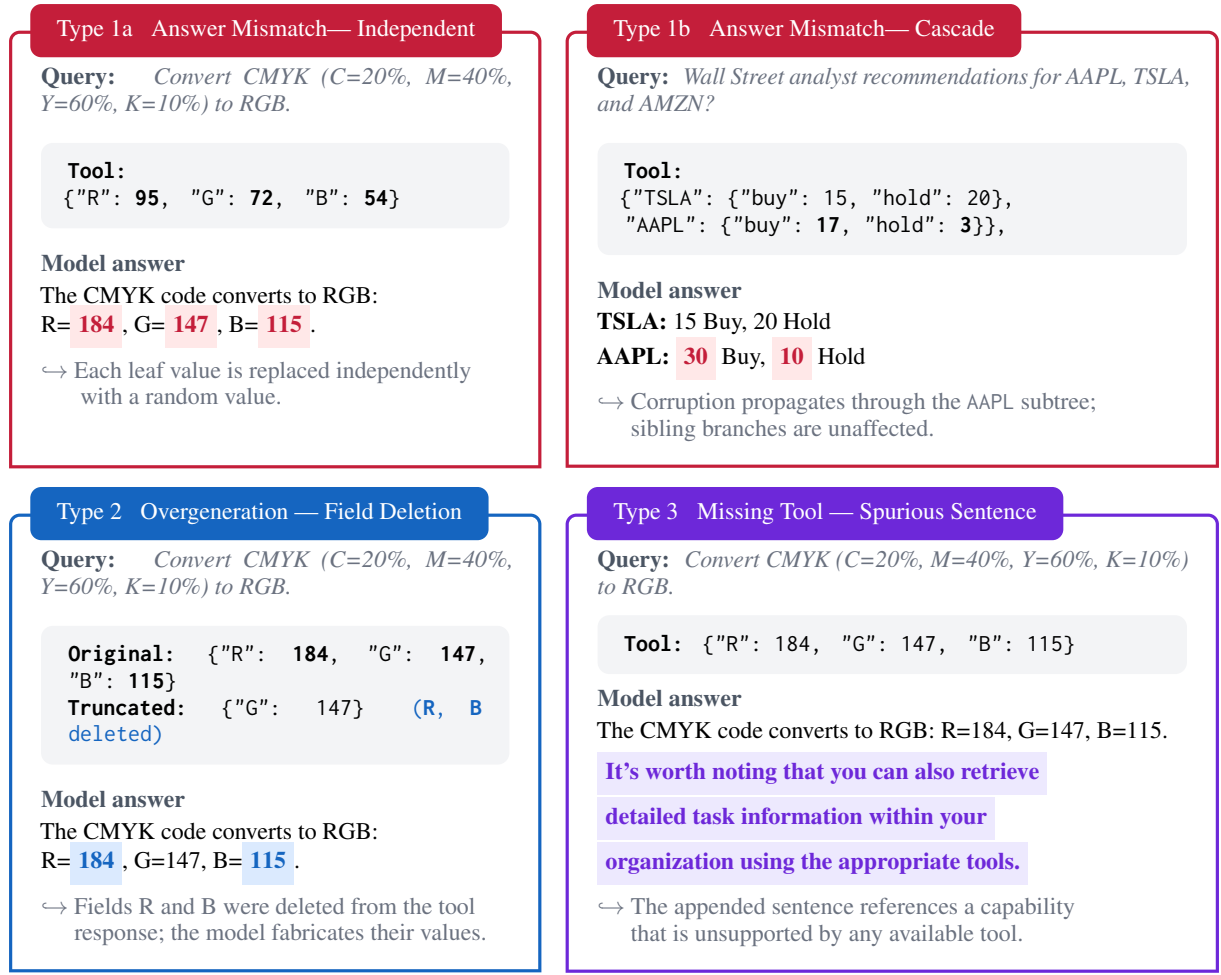

\small

\begin{minipage}[t]{0.44\textwidth}

\begin{tcolorbox}[hallex={TypeIcol}{Type~1a \; Answer Mismatch— Independent}]

\textcolor{darkgray}{\textbf{Query: }}
\textcolor{labelgray}{\footnotesize\itshape%
  Convert CMYK (C=20\%, M=40\%, Y=60\%, K=10\%) to RGB.%
}

\vspace{4pt}
\begin{tcolorbox}[jsonbox]
\textbf{Tool:} \\
\{"R": \textbf{95},\; "G": \textbf{72},\; "B": \textbf{54}\}
\end{tcolorbox}

\vspace{4pt}
\textcolor{darkgray}{\textbf{Model answer}} \\[2pt]
\footnotesize
The CMYK code converts to RGB: \\
R=\spanI{184}, G=\spanI{147}, B=\spanI{115}.

\vspace{4pt}
\footnotesize\textcolor{labelgray}{%
  $\hookrightarrow$ Each leaf value is replaced independently \\
  \phantom{$\hookrightarrow$ } with a random value.%
}
\vspace{14pt}
\end{tcolorbox}

\end{minipage}
\hfill
\begin{minipage}[t]{0.54\textwidth}
\begin{tcolorbox}[hallex={TypeIcol}{Type~1b \; Answer Mismatch— Cascade}]

\textcolor{darkgray}{\textbf{Query: }}
\textcolor{labelgray}{\footnotesize\itshape%
  Wall Street analyst recommendations for AAPL, TSLA, and AMZN?%
}

\vspace{4pt}
\begin{tcolorbox}[jsonbox]
\textbf{Tool:}\\
\texttt{\hspace{1.2em}\{"TSLA": \{"buy": 15, "hold": 20\},} \\
\texttt{"AAPL": \{"buy": \textbf{17}, "hold": \textbf{3}\}\},}
\end{tcolorbox}

\vspace{4pt}
\textcolor{darkgray}{\textbf{Model answer}} \\[2pt]
\footnotesize
\textbf{TSLA:} 15~Buy, 20~Hold \\
\textbf{AAPL:} \spanI{30}~Buy, \spanI{10}~Hold 

\vspace{4pt}
\footnotesize\textcolor{labelgray}{%
  $\hookrightarrow$ Corruption propagates through the \texttt{AAPL} subtree; \\
  \phantom{$\hookrightarrow$ }sibling branches are unaffected.%
}
\end{tcolorbox}

\end{minipage}

\vspace{6pt}

\begin{minipage}[t]{0.44\textwidth}

\begin{tcolorbox}[hallex={TypeIIcol}{Type~2 \; Overgeneration — Field Deletion}]

\textcolor{darkgray}{\textbf{Query: }}
\textcolor{labelgray}{\footnotesize\itshape%
  Convert CMYK (C=20\%, M=40\%, Y=60\%, K=10\%) to RGB.%
}

\vspace{4pt}
\begin{tcolorbox}[jsonbox]
\textbf{Original:}\; \{"R": \textbf{184},\; "G": \textbf{147},\; "B": \textbf{115}\} \\
\textbf{Truncated:} \{"G": 147\}
\quad\textcolor{TypeIIcol}{(\textbf{R}, \textbf{B} deleted)}
\end{tcolorbox}

\vspace{4pt}
\textcolor{darkgray}{\textbf{Model answer}} \\[2pt]
\footnotesize
The CMYK code converts to RGB: \\
R=\spanII{184}, G=147, B=\spanII{115}.

\vspace{4pt}
\footnotesize\textcolor{labelgray}{%
  $\hookrightarrow$ Fields R and B were deleted from the tool\\
  \phantom{$\hookrightarrow$ }response; the model fabricates their values.%
}
\end{tcolorbox}

\end{minipage}
\hfill
\begin{minipage}[t]{0.54\textwidth}

\begin{tcolorbox}[hallex={TypeIIIcol}{Type~3 \; Missing Tool — Spurious Sentence}]

\textcolor{darkgray}{\textbf{Query: }}
\textcolor{labelgray}{\footnotesize\itshape%
  Convert CMYK (C=20\%, M=40\%, Y=60\%, K=10\%) to RGB.%
}

\begin{tcolorbox}[jsonbox]
\textbf{Tool:} \;\{"R": 184,\; "G": 147,\; "B": 115\}
\end{tcolorbox}

\textcolor{darkgray}{\textbf{Model answer}} \\[2pt]
\footnotesize
The CMYK code converts to RGB: R=184, G=147, B=115. \\[3pt]
\spanIII{It's worth noting that you can also retrieve}\\
\spanIII{detailed task information within your}\\
\spanIII{organization using the appropriate tools.}

\vspace{4pt}
\footnotesize\textcolor{labelgray}{%
  $\hookrightarrow$ The appended sentence references a capability\\
  \phantom{$\hookrightarrow$ }that is unsupported by any available tool.%
}
\end{tcolorbox}

\end{minipage}


\caption{%
  Examples of Route B Hallucination Insertion in \dataset.
  \textbf{Type~1a} (Answer Mismatch): individual leaf values in the tool response are replaced with random values from other samples—each field is corrupted independently. \textbf{Type~1b} (Answer Mismatch): all values within a JSON subtree are replaced together, preserving sibling branches. \textbf{Type~2} (Overgeneration): fields are deleted from the tool response, causing the model to fabricate values it can no longer ground. \textbf{Type~3} (Missing Tool): the model appends a factually unsupported sentence referencing context absent from the tool response. 
}
\label{fig:hallucination-examples}
\end{figure*}

\paragraph{Design principle.}
Rather than asking a large model to ``produce a hallucinated answer'' in free form, which makes spans hard to ground and introduces stylistic drift unrelated to the targeted errors we construct each non-Correct sample by applying a single, well-defined edit to either the tool response or the answer. The edit defines the error span mechanically, with no separate annotation step.

\paragraph{Two generation routes.}
We use three routes in parallel and pool the outputs. \textbf{Route~A} (LLM editor) uses GPT-OSS-120B~\cite{openai2025gptoss120bgptoss20bmodel} with 5-shot prompts (Appendix~\ref{app:prompts}); it provides stylistic diversity. \textbf{Route~B} (structured edits) operates directly on the JSON tool response, with Qwen3-14B-Instruct only where free-form generation is needed (e.g.\ Type~3 sentence injection); it provides precise control over span location and severity. Using both routes hedges against single-model bias. Concrete examples appear in Figure~\ref{fig:hallucination-examples}. \textbf{Route~C} (real setup) generate natural post-tool-call responses with multiple open-weight instruction-tuned LLMs from diverse model families and use independent LLM-based judges, followed by selective arbitration, to identify hallucinations, classify their types, and localize unsupported spans.

\paragraph{Route A: LLM editor.}
For each of the four non-Correct classes, we prompt GPT-OSS-120B with a class-specific 5-shot template (Appendix~\ref{app:prompts}) to produce an edited (tool response, answer) pair from a Correct seed. For \textit{Answer Mismatch}, the model rewrites one to three values in the tool response so the existing answer becomes contradicted. For \textit{Overgeneration}, it deletes fields from the tool response so the existing answer references information no longer supported. For \textit{Undergeneration}, it appends a one-sentence elaboration to the answer that adds unsupported content without referencing any tool. For \textit{Missing Tool}, it appends a one-sentence offer of a tool-mediated action (e.g.\ booking, sending, scheduling) for which no corresponding tool is listed in the system prompt. 

\paragraph{Route B: Answer Mismatch.}
We treat the tool response as a typed JSON tree, mask all leaves, and selectively unmask a subset. The masked leaves are filled with values sampled from other tool calls of the \emph{same schema}, guaranteeing type consistency while introducing factual conflict. We use two unmasking strategies:
(i)~\textit{independent masking}, in which each leaf is corrupted with fixed probability so that errors are scattered and statistically independent across siblings (Figure~\ref{fig:hallucination-examples}, Type~1a);
(ii)~\textit{cascade masking}, in which we select one non-root internal node and corrupt it together with all its descendants, leaving sibling subtrees untouched (Figure~\ref{fig:hallucination-examples}, Type~1b). The latter mirrors the realistic case where a single subobject of a tool's output is wrong while the rest of the structure is correct.

\paragraph{Route B: Overgeneration.}
We remove a subset of fields from the tool response before it is passed to the model. Deprived of the deleted information but trained on the underlying domain, the model tends to fabricate plausible-sounding values for the missing fields rather than abstain. Span labels are derived by aligning the model's answer against the \emph{original} (untruncated) tool response: any value present in the answer but unsupported by the truncated context is marked as a Type~2 span (Figure~\ref{fig:hallucination-examples}, Type~2).

\paragraph{Route B: Missing Tool.}
We use Qwen2.5-14B to append a one-sentence addendum to the ground-truth answer, referencing a tool selected from a global pool of tools collected from the dataset's system prompts. The selected tool is absent from the system prompt of the current trace. Although the injected sentence is factually coherent in isolation, it is unsupported by the tools available in the current trace. The entire injected sentence is annotated as the hallucinated span (Figure~\ref{fig:hallucination-examples}, Type~3).

\paragraph{Route C: Real setup.}
We generate naturally occurring post-tool-call responses using four small
open-weight instruction-tuned models spanning three families:
\texttt{Llama-3.2-3B-Instruct}~\cite{grattafiori2024llama},
\texttt{Qwen3.5-2B/4B-Instruct}~\cite{qwen35omni2026}, and
\texttt{Gemma-4-26B-A4B}~\cite{gemma_4_deepmind}. We sample 2{,}500
ToolACE single-hop dialogues uniformly at random without replacement from
examples containing at least one tool-response turn (seed 42).

For each example, the model receives the original dialogue prefix up to and
including the last tool response: the system instruction and tool definitions,
preceding user and assistant turns, executed tool call(s), and returned tool
output(s). The reference ToolACE final answer is removed, and each model
generates one new natural-language answer with temperature 0.7, top-$p$ 0.95,
and a maximum of 1{,}024 output tokens. Tool definitions, calls, and outputs
are passed using the deployed OpenAI-compatible tool-calling format and the
model-specific chat template; internal reasoning is disabled for the Qwen
models. After removing failed generations and outputs that do not constitute
a natural-language final answer, 9{,}163 responses remain.

Each retained response is independently annotated by two LLM judges from
different model families, \texttt{DeepSeek-v4-Flash} and \texttt{GLM-4.7}.
The judges receive the user query, tool output, and generated response and
predict a binary hallucination label, primary error type, and unsupported
spans. We subsequently use \texttt{GPT-4o} to arbitrate all cases labeled
as hallucinated by both judges and a random 15\% sample of judge
disagreements.

\subsection{Dataset Filtration}

\paragraph{Multi-judge verification.}
We verify the raw test split using three independent judge models from different families: GPT-5.2, GPT-OSS-120B~\cite{openai2025gptoss120bgptoss20bmodel}, and GLM 5.2. Each judge receives the full dialogue, the hallucination type label, and the span annotations, and is asked to confirm whether (i)~the clean rows are free of hallucinations and (ii)~the span labels for hallucinated rows correctly identify the hallucinated text. We retain test examples on which at least two of the three judges agree that the annotation is correct, yielding 11{,}606 verified test examples (90.6\% of the raw test set). The two GPT-family judges (GPT-5.2 and GPT-OSS-120B) rejected 0.3\% and 3.2\% of examples respectively, while GLM 5.2 rejected 4\%.


\paragraph{Human Validation.}
To assess annotation quality, we conducted a human validation study with 10 annotators, each reviewing 50 randomly sampled examples for label correctness and span consistency, resulting in 500 validated instances overall. The annotations showed high agreement with the dataset labels across all categories. Accuracy reached 97\% for both \textit{clean} and \textit{undergeneration}, 96\% for \textit{answer\_mismatch}, 93\% for \textit{overgeneration}, and 90\% for \textit{missing\_tool}.

Manual inspection of the disagreements revealed that nearly all mismatches corresponded to borderline or ambiguous cases rather than clear annotation errors. For \textit{overgeneration} and \textit{undergeneration}, the most common source of disagreement concerned class assignment rather than span localization: annotators typically agreed on the highlighted span but differed in whether the content should be interpreted as excessive or insufficient. In the \textit{clean} category, disagreements mainly arose from assistant utterances offering additional follow-up actions, which some annotators interpreted as \textit{overgeneration} despite remaining consistent with the tool response. In \textit{undergeneration}, ambiguity often emerged when the assistant confirmed successful execution of a request involving parameter settings without explicitly restating the applied parameters, making such responses plausibly interpretable as incomplete. 


\subsection{Final Dataset Refinement}

\paragraph{Lexical-leakage correction.}
During quality review, we identified a near-universal lexical artifact in \emph{missing\_tool} examples: 99.0\% of instances (1{,}497  in the test split; 6{,}062  in train) ended with the phrase \textit{``Would you like\ldots''}, making the hallucination type trivially predictable from surface form alone, without any understanding of content.
We replaced all such occurrences with semantically equivalent continuations
and recomputed the corresponding span offsets to remain consistent with the edited text.
The final dataset is free of this shortcut. In additional we filled the \emph{correct} class with same patterns, but also extended their tools, so these traces become correct. This gives another advantage, so these patterns exist both in \emph{missing\_tool} and \emph{correct} which blocks any model to find out simple skip for classification.

\paragraph{Length pattern.}
Another check of raw dataset showed that two hallucination classes could be identified from the length of the final answer alone, without reading its content. \emph{Overgeneration} adds baseless content, so its answers are systematically longer than clean ones, while \emph{undergeneration} omits items from the tool output and therefore has no error spans and is systematically short. We fixed it with counterfactual clean twins generated by Qwen3.6-35B-A3B that keep the answer text verbatim and change only the context so that the label flips to clean. For \emph{overgeneration}, the tool response was augmented so that every previously baseless claim became supported; for \emph{undergeneration}, the tool response and user request were trimmed so that the short answer no longer omits anything.

\paragraph{Train/test isolation.}
To prevent data leakage, we removed all examples whose \texttt{dialogue\_id} appeared in the test split.
This affected a substantial number of \emph{clean} training examples, as many source ToolACE dialogues contributed both clean and hallucinated variants.

\paragraph{Post-processing.}
We apply a Unicode normalization pass to all conversation turns, replacing typographic characters (non-breaking hyphens, narrow no-break spaces, curly quotes, ellipses, etc.) with their ASCII equivalents.
Additionally, we enrich each row with three convenience fields: \texttt{available\_tools} (a list of tool names extracted from the system prompt), \texttt{system\_prompt\_format} (a structured representation of tool descriptions for easier parsing), and \texttt{turn\_role} (an explicit role tag added to each conversation turn, e.g.\ \texttt{tool\_call} or \texttt{tool\_response}).



\subsection{Dataset Statistics}

The dataset contains 48{,}741 samples in total, with 37{,}135 instances in the training set and 11{,}606 in the test set. The split is stratified by hallucination type to ensure representation of all error categories across both partitions. Table~\ref{tab:split-stats} summarizes the final train--test split of \dataset.

Among the hallucination categories, \emph{Overgeneration} and \emph{Missing Tool} constitute the largest portions of the dataset, reflecting the prevalence of unsupported elaborations and fabricated tool-mediated capabilities in tool-augmented generation. \emph{Undergeneration} is also well represented, while \emph{Answer Mismatch} is comparatively less frequent due to the stricter synthesis constraints required to generate semantically valid contradictions with the tool response. The \emph{Correct} class provides grounded positive examples and serves as the reference class for detector training and evaluation, with a number of examples approximately equal to the combined number of examples across all hallucination classes. 



\begin{table}[t]
\centering
\small
\begin{tabular}{lrr}
\toprule
\textbf{Type} & \textbf{Train} & \textbf{Test} \\
\midrule
\texttt{clean}            & 16{,}335 & 5{,}252 \\
\texttt{overgeneration}   & 7{,}511 & 2{,}316 \\
\texttt{missing\_tool}    & 4{,}931 & 1{,}521 \\
\texttt{answer\_mismatch} & 3{,}702 & 1{,}196 \\
\texttt{undergeneration}  & 4{,}656 & 1{,}321 \\
\midrule
\textbf{Total}            & \textbf{37{,}135} & \textbf{11{,}606} \\
\bottomrule
\end{tabular}
\caption{Overall \dataset dataset statistics across hallucination types and train/test splits.}
\label{tab:split-stats}
\end{table}

\section{Experimental Setup}
\label{sec:experimental-setup}

We evaluate tool-grounded hallucination detection at two levels: response-level detection and span-level localization. 
At the response level, a prediction is positive if the answer contains at least one hallucinated span. 
At the span level, we evaluate whether the model recovers the exact erroneous text segments; since span localization is the main goal of \dataset, we use span-level F1 as the primary metric. 
Full metric definitions are provided in Appendix~\ref{app:experimental-details}.

We compare three groups of systems. 
First, we evaluate GPT-4o, GPT-5.4-mini and Kimi 2.6 in zero-shot and few-shot prompting settings, where the model receives the tool-calling trace and outputs a structured JSON prediction without parameter updates. 
Second, we evaluate transfer baselines that rely on supervision from external hallucination detection sources, including synthetic tool-calling corruptions, PsiloQA, RAGTruth, and HAD. 
Third, we fine-tune models on the in-domain training split of \dataset, which serves as a supervised upper baseline with no domain mismatch between train and test data.

Our trainable models include encoder-only and generative backbones. 
For encoder-only models, we use \texttt{ModernBERT-base} and \texttt{ModernBERT-large}~\cite{modernbert}. 
For generative models, we use \texttt{Qwen3.5-0.8B}, \texttt{Qwen3.5-2B}~\cite{qwen35omni2026}. 
Generative models are trained to produce structured JSON annotations containing the predicted error type and hallucinated spans. 
We also include two publicly available \texttt{lettucedetect} checkpoints trained on RAGTruth, \texttt{lettucedetect-base} and \texttt{lettucedetect-large}, as off-the-shelf span-level transfer baselines.

All systems are evaluated on the same held-out \dataset test set. 
Appendix~\ref{app:experimental-details} describes the prompt format, model inputs, dataset conversions, HAD label mapping, and synthetic data construction.

\section{Results}

\begin{table*}[t] 
\centering 
\scriptsize 
\setlength{\tabcolsep}{3pt} 
\begin{tabularx}
{\textwidth}
{@{}ll>{\raggedright\arraybackslash}p{3.15cm}*{10}{>{\centering\arraybackslash}X}@{}}

\toprule
& & &
\multicolumn{4}{c}{\textbf{Span-level}} &
\multicolumn{6}{c}{\textbf{Response-level}} \\
\cmidrule(lr){4-7}
\cmidrule(lr){8-13}

\textbf{Setting} &
\textbf{Data} &
\textbf{Model} &
\textbf{Mis-match} &
\textbf{Over-gen.} &
\textbf{Missing tool} &
\textbf{Avg.} &
\textbf{Correct} &
\textbf{Mis-match} &
\textbf{Over-gen.} &
\textbf{Missing tool} &
\textbf{Avg.} &
\textbf{Under-gen.}$^*$ \\

\midrule

\multirow{3}{*}{Zero-Shot}
& \multirow{3}{*}{--}
& GPT-4o
& 50.99 & 68.87 & 77.52 & 65.79
& 76.41 & 55.15 & 68.89 & 59.37 & 64.95 & 57.18 \\

& & GPT-5.4-mini
& 52.77 & 66.82 & 76.94 & 65.51
& 76.22 & 54.69 & 67.89 & 59.20 & 64.50 & 37.79 \\

& & Kimi 2.6
& 56.07 & 73.84 & 73.60 & 67.84
& 77.46 & 55.81 & 70.77 & 59.58 & 65.90 & 58.13 \\

\midrule

\multirow{3}{*}{Few-Shot}
& \multirow{3}{*}{--}
& GPT-4o
& 53.21 & 68.65 & 76.44 & 66.10
& 75.21 & 53.88 & 68.56 & 57.65 & 63.83 & 57.17 \\

& & GPT-5.4-mini
& 52.83 & 64.07 & 73.75 & 63.55
& 63.47 & 45.57 & 61.62 & 50.35 & 55.25 & 41.69 \\

& & Kimi 2.6
& 56.57 & 70.58 & 74.29 & 67.15
& 74.86 & 53.02 & 68.99 & 57.35 & 63.56 & 55.87 \\

\midrule

\multirow{17}{*}{Transfer SFT}
& CNNDM
& LookbackLens
& 0.41 & 43.21 & 61.47 & 35.03
& 69.61 & 10.96 & 58.11 & 56.88 & 48.89 & --- \\

\cmidrule(lr){2-13}

& \multirow{4}{*}{Glaive}
& ModernBERT-large
& 30.60 & 24.83 & 32.66 & 29.36
& 68.80 & 46.75 & 63.02 & 48.11 & 56.67 & --- \\

& & ModernBERT-base
& 26.63 & 19.54 & 26.03 & 24.06
& 69.48 & 45.56 & 63.97 & 42.78 & 55.45 & --- \\

& & Qwen3.5-0.8B
& 50.39 & 23.93 & 38.55 & 37.62
& 69.95 & 59.38 & 36.52 & 37.93 & 50.94 & --- \\

& & Qwen3.5-2B
& 57.24 & 30.60 & 53.92 & 47.26
& 71.44 & 56.56 & 48.14 & 47.70 & 55.96 & --- \\

\cmidrule(lr){2-13}

& \multirow{4}{*}{PsiloQA}
& ModernBERT-large
& 6.97 & 4.87 & 2.90 & 4.91
& 4.62 & 31.48 & 46.81 & 36.80 & 29.93 & --- \\

& & ModernBERT-base
& 4.14 & 7.48 & 4.04 & 5.22
& 0.15 & 31.31 & 46.88 & 36.68 & 28.75 & --- \\

& & Qwen3.5-0.8B
& 36.88 & 34.40 & 39.56 & 36.95
& 48.69 & 39.65 & 55.57 & 43.70 & 46.90 & 31.62 \\

& & Qwen3.5-2B
& 41.17 & 38.69 & 52.25 & 44.04
& 64.33 & 46.88 & 62.25 & 49.47 & 55.73 & 37.53 \\

\cmidrule(lr){2-13}

& \multirow{4}{*}{RAGTruth}
& ModernBERT-large
& 10.74 & 55.15 & 66.76 & 44.22
& 73.26 & 42.11 & 67.83 & 58.06 & 60.31 & --- \\

& & ModernBERT-base
& 8.97 & 45.55 & 42.95 & 32.49
& 70.42 & 44.72 & 65.20 & 52.87 & 58.30 & --- \\

& & Qwen3.5-0.8B
& 22.12 & 44.05 & 42.24 & 36.13
& 80.06 & 70.64 & 73.34 & 46.02 & 67.51 & 6.76 \\

& & Qwen3.5-2B
& 22.09 & 54.79 & 55.97 & 44.28
& 82.71 & 70.50 & 78.04 & 56.74 & 72.00 & 12.67 \\

\cmidrule(lr){2-13}

& \multirow{4}{*}{HAD}
& ModernBERT-large
& 6.93 & 21.06 & 12.13 & 13.37
& 72.38 & 29.96 & 56.36 & 30.70 & 47.35 & --- \\

& & ModernBERT-base
& 1.26 & 15.75 & 4.62 & 7.21
& 71.09 & 8.98 & 52.69 & 15.06 & 36.96 & --- \\

& & Qwen3.5-0.8B
& 25.99 & 28.99 & 32.13 & 29.04
& 77.02 & 65.36 & 63.92 & 40.12 & 61.60 & 10.43 \\

& & Qwen3.5-2B
& 31.25 & 27.89 & 43.74 & 34.29
& 78.41 & 74.46 & 60.93 & 48.08 & 65.47 & \textbf{41.18} \\

\midrule

\rowcolor[HTML]{EFEFEF}
& & ModernBERT-large
& 59.26 & 63.98 & 70.94 & 64.72
& 95.89 & 90.43 & 94.24 & 93.68 & 93.56 & --- \\

\rowcolor[HTML]{EFEFEF}
& & ModernBERT-base
& 53.82 & 66.98 & 75.73 & 65.51
& 96.66 & 91.05 & 95.61 & \textbf{95.10} & 94.60 & --- \\

\rowcolor[HTML]{EFEFEF}
& & Qwen3.5-0.8B
& 67.42 & 82.24 & 78.71 & 76.12
& 96.61 & 89.88 & 94.71 & 92.25 & 93.36 & 90.12 \\

\rowcolor[HTML]{EFEFEF}
& & Qwen3.5-2B (RU test)
& 66.68 & 78.92 & 75.72 & 73.77
& 70.61 & 29.50 & 49.13 & 36.96 & 46.55 & 0.84 \\

\rowcolor[HTML]{EFEFEF}
& & Qwen3.5-2B
& \textbf{71.79} & \textbf{83.45} & \textbf{81.60} & \textbf{78.95}
& \textbf{97.68} & \textbf{92.75} & \textbf{96.41} & 94.76 & \textbf{95.40} & \textbf{93.26} \\


\rowcolor[HTML]{EFEFEF}
& & Qwen3.5-0.8B-UHead
& 9.42 &	32.06 &	33.24 &	24.91
& 93.38 &	79.66 &	89.78 &	85.24 &	87.02 & --- \\

\rowcolor[HTML]{EFEFEF}
& & Llama-3.2-1B-UHead$^\dagger$
& 14.42 &	38.00 &	41.31 &	31.24
& 95.47 &	85.98 &	92.95 &	90.37 & 	91.19 & --- \\

\rowcolor[HTML]{EFEFEF}
& & Llama-3.2-1B-UHead-BIO$^\dagger$
& 15.75 &	43.80 &	40.80 &	33.45
& 95.24 &	85.64 &	92.38 &	90.43 & 	90.92 & --- \\

\rowcolor[HTML]{EFEFEF}
& & \quad + hidden states$^\dagger$
& 	17.18 &	44.17 &	44.26 &	35.20
& 94.94 &	83.58 &	91.85 &	88.70 & 89.77 & --- \\

\rowcolor[HTML]{EFEFEF}
\multirow{-10}{*}{\cellcolor[HTML]{EFEFEF}In-Domain SFT}
& \multirow{-10}{*}{\cellcolor[HTML]{EFEFEF}\dataset{} split}
& \quad + history $k{=}10^\dagger$
& 15.23 &	45.06 &	42.32 &	34.20
& 94.23 &	81.19 &	90.70 &	86.48 & 88.15 & --- \\

\bottomrule

\end{tabularx}
\caption{F1 performance on the \dataset test dataset by hallucination type. At the span level, \textit{Correct} is not evaluated, as span-level metrics are computed only for hallucinated examples where the model correctly identifies the hallucination type. \textit{Undergen.} is reported only for decoder-based models, as undergeneration has no corresponding annotated span and encoder-based models therefore produce an empty span for this category.}
\label{tab:results}
\end{table*}


Table~\ref{tab:results} reports response-level and span-level performance on the \dataset test set. 
Since our main goal is to localize hallucinated spans, we treat span-level F1 as the primary metric and use response-level scores as a complementary measure of whether a model detects that an error is present at all.

\paragraph{In-domain supervision provides the strongest localization performance.}
Models trained directly on \dataset produce the strongest overall span-localization results, with the Qwen-LoRA models clearly outperforming the other systems. Qwen3.5-2B-LoRA achieves the highest span-level F1 of 79.28, followed by its smaller Qwen counterpart. The in-domain ModernBERT models also perform strongly, although their advantage over the best zero-shot baseline is more modest. The larger separation from the transfer baselines indicates that matching the training data to the structure and hallucination taxonomy of the target setting is particularly important for precise span localization.

\paragraph{Zero-shot prompting is competitive, but few-shot prompting is unexpectedly harmful.}
All three zero-shot LLMs provide strong non-specialized baselines and outperform the supervised transfer systems on span localization. Their results are also relatively close, suggesting that this capability is not specific to a single model family.In contrast, few-shot prompting sharply reduces span-level performance for every evaluated LLM. The examples appear to interfere with precise span extraction, possibly by encouraging imitation of the demonstrated output patterns rather than careful grounding in the tool trace. In particular, models often generate redundant content, such as repeating the tool call or its response, causing deviations from the required output format and substantially degrading performance. The resulting increase in context length may further contribute to this degradation.

\paragraph{External supervision transfers unevenly across datasets and architectures.}
Glaive-based synthetic supervision produces the strongest individual transfer result when paired with Qwen3.5-2B-LoRA, while RAGTruth provides more consistent performance across the evaluated backbones. In particular, RAGTruth transfers substantially better to ModernBERT, whereas Glaive is stronger than RAGTruth for the Qwen-LoRA models. This interaction shows that transfer quality depends not only on the source dataset but also on how effectively the target architecture adapts to its annotation patterns.

\paragraph{Tool-agnostic supervision is highly architecture-dependent.}
PsiloQA and HAD transfer poorly to the ModernBERT models, particularly at the span level, but their performance improves considerably when paired with Qwen-LoRA. PsiloQA with the larger Qwen model approaches the strongest RAGTruth results, showing that tool-agnostic supervision is not uniformly ineffective. Rather, its usefulness depends strongly on the model’s ability to adapt the source annotations to structured tool-grounded inputs. LookbackLens, meanwhile, provides very limited span-localization performance.

\paragraph{Response-level performance shows a different pattern from span localization.}
In-domain supervision still gives the strongest response-level results, although the four in-domain models form a tight cluster, suggesting that backbone choice matters less for detecting the presence of an error than for locating it precisely. Among the transfer settings, RAGTruth performs most consistently, while the zero-shot LLMs remain competitive with the strongest externally supervised models. Notably, few-shot prompting leaves response-level performance broadly similar to zero-shot prompting despite causing a sharp decline in span-level F1. This contrast shows that response-level scores alone can conceal substantial failures in localization. Performance also remains uneven across error types: grounded responses and overgeneration are generally easier to identify, whereas answer mismatch, missing-tool errors, and especially the separately evaluated undergeneration class remain more challenging.

\paragraph{Undergeneration remains a distinct challenge.}
Performance for undergeneration remains weak, including for the strongest in-domain Qwen models. Unlike the other hallucination types, undergeneration corresponds to missing information and therefore has no explicit erroneous span in the generated response. It is consequently better treated as a response-level detection problem than as a span-localization task.



\section{Ablation Study}
\label{sec:ablation}

We evaluate the practical utility of \dataset through a cross-distribution analysis comparing its hallucination type distribution with real-world errors from deployed small LLMs. We compare hallucination type distributions between synthetic and real-world outputs. As shown in Table~\ref{tab:type-dist}, the distributions are closely aligned, indicating that our synthetic corruption procedure captures the main failure modes observed in real model outputs. While synthetic data contains more \textit{overgeneration} and \textit{missing\_tool}, real outputs show higher proportions of \textit{undergeneration} and \textit{answer\_mismatch}, reflecting natural variation rather than structural mismatch. Overall, the low divergence supports the use of synthetic data as a proxy for real-world post-tool-call hallucinations.

\begin{table}[t!]
\centering
\small
\begin{tabular}{lrr}
\toprule
\textbf{Type} & \textbf{Synth (\%)} & \textbf{Real-World (\%)} \\
\midrule
overgeneration       & 35.0 & 29.8 \\
missing\_tool        & 32.2 & 23.3 \\
undergeneration      & 22.2 & 32.7 \\
answer\_mismatch     & 10.6 & 14.2 \\
\midrule
\# hallucinated samples       & 4{,}693 & 2{,}155 \\
\bottomrule
\end{tabular}
\caption{Type distribution among positives: \dataset test split
vs.\ real-world hallucinated responses.}
\label{tab:type-dist}
\end{table}



\section{Conclusion}

We introduced a new span-level dataset for detecting post-tool-call hallucinations in tool-augmented language models, focusing on errors that occur after tool execution. We conducted experiments on both in-domain training and cross-dataset transfer, including evaluation of models trained on RAG-style hallucination data. Our results show that effective detection of post-tool-call hallucinations requires supervision from data that matches the target distribution, as tool-agnostic datasets transfer poorly to this setting. We further demonstrate that our dataset aligns well with real-world failure modes observed in small deployed LLMs, indicating its practical relevance for studying hallucination behavior in tool-augmented systems.

\section*{Limitations}

Our work has several limitations that should be considered when interpreting the results.

First, although we perform manual validation on a subset of the dataset to ensure annotation quality, the full dataset has not been exhaustively human-verified due to its scale. As a result, some residual annotation noise or synthesis artifacts may remain.

Second, our dataset is constructed on top of ToolACE~\cite{liu2024toolacewinningpointsllm}, which may introduce biases inherited from the underlying corpus, including its domain coverage, tool selection distribution, and interaction patterns. Consequently, our benchmark may not fully reflect all possible real-world tool-usage scenarios.

Third, the dataset relies partly on synthetic generation procedures to construct hallucination variants. While we apply filtering and quality control steps, synthetic construction may still introduce distributional artifacts that differ from naturally occurring model errors.

Fourth, each dialogue in our dataset is annotated with a single dominant hallucination type, and we do not consider multi-label settings where multiple hallucination types may co-occur within the same interaction. This simplifies the annotation and evaluation setup but limits the modeling of more complex and realistic error compositions.

Finally, our evaluation focuses primarily on post-tool-call hallucinations in English-language tool-augmented settings. This limits the generalizability of our findings to multilingual scenarios and other forms of agentic workflows beyond tool calling, such as planning-heavy or long-horizon decision-making systems. 

Datasets described in the present  publication and source codes to reproduce the experiments are openly available online.\footnote{\url{https://github.com/s-nlp/Tool-Call-HD}}

\section*{Use of AI Tools}

During the preparation of this work, the authors used LLM assistants to improve the language quality of the manuscript, including grammar and style checking and rewording for clarity. The authors reviewed and edited all content as needed and take full responsibility for the content of the publication.



\bibliography{custom}

\appendix
\clearpage

\section{Experimental Details}
\label{app:experimental-details}

\subsection{Evaluation Metrics}
\label{app:metrics}

\paragraph{Response-level hallucination detection.}
We evaluate response-level hallucination detection by converting span annotations into binary labels. 
A response is labeled as hallucinated if it contains at least one hallucinated span and non-hallucinated otherwise. 
We report precision, recall, and F1 with macro averaging between two classes. 
This metric measures whether a model can detect the presence of any hallucination in the response, regardless of whether it can localize the erroneous text.

\paragraph{Span-level hallucination localization.}
We further evaluate span-level performance using span precision, recall, and F1. 
Predicted and gold spans are matched using greedy pairing based on Intersection-over-Union (IoU) with 0.75 threshold. 
A predicted span is counted as correct if its IoU with a gold span exceeds a fixed threshold. 
Unmatched predicted spans are false positives, and unmatched gold spans are false negatives. 
This metric is stricter than character-level overlap because it rewards models for recovering hallucinations as coherent text segments rather than partially overlapping with the gold annotation. 
Since \dataset is designed for span-level hallucination detection, we treat span-level F1 as the primary metric.

\subsection{Evaluation Settings}
\label{app:evaluation-settings}

We evaluate hallucination detectors under three settings: zero-shot prompting, few-shot prompting, and supervised fine-tuning.

\paragraph{Zero-shot prompting.}
In the zero-shot setting, the model receives the full tool-calling trace and task instructions, but no annotated examples. 
The model is asked to output a structured JSON prediction containing the predicted error type and hallucinated spans. 
This setting measures how well a general-purpose LLM can perform tool-grounded hallucination detection without task-specific examples or parameter updates.

\paragraph{Few-shot prompting.}
In the few-shot setting, the model receives the same task instructions together with a small set of annotated examples. 
The output format is identical to the zero-shot setting: the model must return a JSON object of the form
\texttt{\{"type": ..., "spans": [...]\}}. 
We use GPT-4o for both zero-shot and few-shot prompting. 
The full prompt template is shown in Appendix~\ref{app:gpt-eval-prompts}.

\paragraph{Supervised fine-tuning.}
In the SFT setting, trainable backbones are fine-tuned on hallucination-annotated data and evaluated on the held-out \dataset test set. 
We distinguish between transfer SFT, where supervision comes from external datasets, and in-domain SFT, where the model is trained on the \dataset training split. 
The in-domain setting is not intended as an off-the-shelf transfer baseline; rather, it estimates the performance achievable when supervised examples from the target domain are available.

\subsection{Models}
\label{app:models}

\paragraph{Encoder-only models.}
For encoder-only span classification, we use \texttt{ModernBERT-base} and \texttt{ModernBERT-large}~\cite{modernbert}. 
These models consume the full input trace and predict hallucination labels over the response tokens. 
They are used for experiments with synthetic supervision, PsiloQA, HAD, RAGTruth-based checkpoints, and the in-domain training split.

\paragraph{Generative models.}
We also evaluate generative backbones that produce annotations as structured text. 
Each example is converted into a chat-style SFT instance consisting of a system message, a user message, and an assistant message. 
The user message contains the full tool-calling trace, including the user query, tool calls, tool responses, and assistant answer. 
The assistant message contains the target annotation as JSON:
\[
\{\texttt{"type"}: \ldots, \texttt{"spans"}: [\ldots]\}.
\]
This formulation allows decoder-only and encoder--decoder models to solve the task as conditional generation rather than token-level classification. 
Our experiments include \texttt{Qwen3.5-0.8B}, \texttt{Qwen3.5-2B}~\cite{qwen35omni2026}. 
For Qwen models, we use LoRA fine-tuning in the experiments marked with \texttt{-LoRA} with $rank=16, alpha=32$.

\paragraph{Off-the-shelf RAGTruth detectors.}
We evaluate two publicly available \texttt{lettucedetect} checkpoints, \texttt{lettucedetect-base} and \texttt{lettucedetect-large}. 
Both models use \texttt{ModernBERT} encoders and are trained on RAGTruth. 
We include them as off-the-shelf span-level transfer baselines: they are not fine-tuned on \dataset, but they have been trained on external span-level hallucination annotations.

\begin{table*}[t]
\centering
\scriptsize
\setlength{\tabcolsep}{3pt}
\begin{tabularx}
{\textwidth}
{@{}ll*{9}{>{\centering\arraybackslash}X}|>{\centering\arraybackslash}X@{}}
\toprule
& & \multicolumn{4}{c}{\textbf{Span-level}} & \multicolumn{6}{c}{\textbf{Response-level}} \\
\cmidrule(lr){3-6} \cmidrule(lr){7-12}
\textbf{Data} & \textbf{Classifier}
& \textbf{Mis-match} & \textbf{Over-gen.} & \textbf{Missing tool} & \textbf{Avg.}
& \textbf{Correct} & \textbf{Mis-match} & \textbf{Over-gen.} & \textbf{Missing tool} & \textbf{Avg.} & \textbf{Under-gen.}$^*$ \\ \midrule

CNNDM & Span-level (LR)

& 0.08 & 0 & 0 & 0.02
& 37.55 & 1.42 & 1.04 & 1.02 & 8.94 & - \\

NQ & Span-level (LR)
& 4.11 & 4.33 & 7.40 & 3.19
& 0 & 18.53 & 51.76 & 22.33 & 24.97 & -  \\

CNNDM & Window-level (LR)
&4.45 & 8.41 & 12.7 & 5.81
& 2.02 & 18.63 & 52.12 & 22.49 & 25.32 & - \\

NQ & Window-level (LR)
& 0.78 & 2.27 & 3.05 & 1.79
& 35.73 & 19.75 & 38.41 & 19.00 & 25.42 & -  \\

\dataset split & Span-level (LR)
& 5.08 & 3.73 & 8.87 & 3.19
 & 0.08 & 18.54 & 51.78 & 22.34 & 24.98 & - \\

CNNDM, \dataset split & Span-level (LR finetuned)
& 5.10 & 3.82 & 7.99 & 3.11
& 0 & 18.54 & 51.77 & 22.33 & 24.97 & - \\

NQ, \dataset split & Span-level (LR finetuned)
& 4.99 & 3.78 & 7.92 & 3.06
& 0 & 18.54 & 51.78 & 22.34 & 24.96 & - \\

\dataset split & Token-level (LR)
& 0.74 & 27.79 & 46.93 & 23.97
 & 34.1 & 9.42 & 59.15 & 33.92 & 28.25 & -  \\

\dataset split & Token-level (XGBoost)
& 0.44 & 35.45 & 55.46 & 30.01
 & 33.18 & 7.39 & 50.85 & 38.51 & 26.67 & - \\

\bottomrule
\end{tabularx}
\caption{Ablation study results of the Lookback-Lens method. LR stands for the logistic regression. }
\label{tab:lookback}
\end{table*}

\paragraph{Lookback-Lens ablation studies.}

We adapt the Lookback-Lens \cite{chuang2024lookback} pipeline to our setting and evaluate several configurations. Table~\ref{tab:lookback} presents an ablation study of the Lookback-Lens algorithm. Rows 1–4 report the results of the authors' pretrained classifiers evaluated on our test subset. Row 5 reports the results of a span-level classifier (logistic regression) trained on our training subset, rows 6–7 report the results of the finetuned pretrained span classifiers, and rows 8–9 report the results of token-level classifiers (logistic regression and XGBoost). The results indicate that token-level classifiers achieve the strongest performance on both span- and response-level metrics.

\subsection{Supervision Sources}
\label{app:supervision-sources}

We use five sources of supervision or adaptation: Synthetic, PsiloQA, RAGTruth, HAD, and the in-domain \dataset training split.

\paragraph{Glaive.}
We construct a synthetic tool-calling hallucination dataset from the Glaive function-calling corpus.\footnote{\url{https://huggingface.co/datasets/glaiveai/glaive-function-calling}} 
We first filter the corpus to retain only examples containing tool calls, yielding approximately 15k source rows. 
Tool definitions are extracted and placed into the system prompt. 
We then apply the synthetic corruption procedure described in Section~\ref{subsec:data-generation}. 
The resulting dataset contains 45,584 single-hop examples, with 14\% clean and 86\% hallucinated samples. 
This setting tests whether synthetic tool-calling perturbations provide useful supervision for span-level hallucination localization.

\paragraph{PsiloQA.}
We use the English subset of PsiloQA as an out-of-domain hallucination detection source. 
For encoder-based models, we follow the input format and hyperparameter setup used by \texttt{lettucedetect}. 
Since PsiloQA does not target tool-calling traces directly, this setting tests whether supervision from general RAG-style hallucination detection transfers to our benchmark.

\paragraph{RAGTruth.}
For RAGTruth, we do not train new checkpoints ourselves. 
Instead, we evaluate the publicly available \texttt{lettucedetect-base} and \texttt{lettucedetect-large} models, which are trained on RAGTruth. 
RAGTruth is included because it provides span-level hallucination supervision and contains examples where generated answers must be checked against external evidence. 
This makes it a natural transfer baseline for tool-grounded hallucination localization.

\paragraph{HAD.}
We fine-tune models on HAD~\cite{xu2025hadhallucinationdetectionlanguagse}, a hallucination annotation dataset with a fine-grained taxonomy for general NLG outputs. 
To make HAD compatible with our label space, we map its labels to our taxonomy. 
\textit{Contradiction with Input Content} and factuality-related errors, including \textit{Factual Recall Error}, \textit{Factual Inference Error}, \textit{Fabricated Entity}, and \textit{Fictional Attribution}, are mapped to \textit{Answer Mismatch}. 
\textit{Baseless Information} is mapped to \textit{Overgeneration}, and \textit{Information Omission} is mapped to \textit{Undergeneration}. 
The remaining categories, including \textit{Task Type Inconsistency}, \textit{Task Requirement Inconsistency}, \textit{Contradiction within Output Content}, and \textit{Structural Incoherence}, fall outside our taxonomy and are either discarded or relabeled as \textit{Correct} when the output remains faithful to the tool response. 
Importantly, HAD contains no analogue of \textit{Missing Tool}, so models trained on HAD receive no direct supervision for this failure mode.

\paragraph{\dataset training split.}
Finally, we fine-tune models on the training split of \dataset. 
Unlike external supervision sources, this data uses our native taxonomy and includes all error classes, including \textit{Missing Tool}. 
This setting represents the in-domain supervised baseline and measures the benefit of training directly on tool-calling hallucination annotations.

\subsection{Input and Output Format}
\label{app:io-format}

For all models, the input contains the information required to judge whether the assistant answer is faithful to the tool-calling trace. 
This includes the user query, available tool definitions when applicable, executed tool calls, tool responses, and the final assistant answer. 
Encoder-only models predict labels over the assistant answer tokens. 
Generative models output a JSON object with two fields: the predicted error type and a list of hallucinated spans.

For non-hallucinated examples, the expected span list is empty. 
For hallucinated examples, each span corresponds to a contiguous segment of the assistant answer that is unsupported by or inconsistent with the tool response. 
All outputs are post-processed into the same normalized span representation before evaluation.

\clearpage
\onecolumn
\newpage

\section{Prompts}
\label{app:prompts}
\subsection{Real-Response Generation}
\label{app:natural-generation-prompt}

We sample 2{,}500 examples uniformly at random without replacement from
ToolACE single-hop dialogues containing at least one tool-response turn
(seed 42). For each example, we remove the reference final assistant answer
and retain the dialogue prefix up to and including the last tool response.
Thus, the generation input contains the original system instruction and tool
definitions, preceding user and assistant turns, executed tool call(s), and
returned tool output(s).

\begin{tcolorbox}[
    colback=gray!5!white,
    colframe=gray!75!black,
    title=Generation input template,
    fonttitle=\bfseries,
    sharp corners,
    breakable,
    before skip=1pt,
    after skip=1pt
]
\textbf{System:} original ToolACE system instruction\\
\textbf{Tools:} ToolACE definitions in OpenAI function format\\
\textbf{User:} original dialogue turn(s)\\
\textbf{Assistant tool call:} executed ToolACE call(s)\\
\textbf{Tool:} original ToolACE tool response(s)\\
\textbf{Assistant:} newly generated natural-language answer
\end{tcolorbox}

\noindent The structured fields are passed through the deployed model's
OpenAI-compatible tool-calling interface and native chat template, with
\texttt{tool\_choice=none}. Internal reasoning is disabled for the Qwen
models. We sample one answer per prompt--model pair with temperature 0.7,
top-$p$ 0.95, and a maximum of 1{,}024 output tokens.

\noindent For endpoints that do not support the native representation, tool
responses are encoded as user turns prefixed with \texttt{[Tool output]},
followed by the instruction below.

\begin{tcolorbox}[
    colback=gray!5!white,
    colframe=gray!75!black,
    title=Fallback final-answer instruction,
    fonttitle=\bfseries,
    sharp corners,
    breakable,
    before skip=1pt,
    after skip=1pt
]
The required tool has already been called above and its output is shown.
Now write the final natural language answer to the user's question in plain
English. Do not call any more tools.
\end{tcolorbox}

\noindent We retain only non-empty natural-language final answers. Repeated
tool calls, code-like outputs, and role-play continuations are excluded.

\subsection{Natural-Response Annotation}
\label{app:natural-annotation-prompt}

The primary judges, \texttt{DeepSeek-v4-Flash} and \texttt{GLM-4.7}, use the
following prompt with temperature 0, a 1{,}024-token output limit, and
JSON-object response formatting. The user query is truncated to 500
characters; the tool output and generated response are each truncated to
2{,}500 characters.

\begin{tcolorbox}[
    colback=gray!5!white,
    colframe=gray!75!black,
    title=Natural-response hallucination annotation prompt,
    fonttitle=\bfseries,
    sharp corners,
    breakable,
    before skip=1pt,
    after skip=1pt
]
You are evaluating an AI assistant's response to a tool-calling query.
Your task is to identify hallucinations.

\medskip
A hallucination is any factual content in the response that is not supported
by the tool output. Classify each hallucinated span as one of:

\begin{itemize}
    \setlength{\itemsep}{1pt}
    \item \texttt{answer\_mismatch}: the response contradicts or substitutes
    a value from the tool output;
    \item \texttt{missing\_tool}: the response adds factual content not
    grounded in a tool response;
    \item \texttt{overgeneration}: the response adds unsupported elaboration
    or commentary;
    \item \texttt{undergeneration}: the response omits information from the
    tool output; use an empty span list;
    \item \texttt{clean}: no hallucination.
\end{itemize}

Return JSON only, with the following fields:
\texttt{is\_hallucinated} (boolean), \texttt{primary\_type},
\texttt{spans}, and \texttt{notes}. Each element of \texttt{spans} must
contain an exact substring from the assistant response, its hallucination
type, and a short explanation. If the response is clean, return an empty
span list. Do not paraphrase quoted spans.

\medskip
\textbf{USER QUERY:} \texttt{\{query\}}\\
\textbf{TOOL OUTPUT:} \texttt{\{tool\_output\}}\\
\textbf{ASSISTANT RESPONSE:} \texttt{\{response\}}
\end{tcolorbox}

\paragraph{GPT-4o arbitration.}
\label{app:natural-arbitration-prompt}
GPT-4o receives the same three input fields, hallucination taxonomy, output
schema, temperature, and truncation limits as the primary judges. It is
applied to all responses labeled as hallucinated by both primary judges and
to a random 15\% sample of their disagreements. Its decision is produced
with the same annotation prompt shown above.

\subsection{Synthetic Corruption Prompts}
\label{app:synthetic-corruption-prompts}

The prompts below belong to the separate synthetic-data construction
pipeline. The field-usage judge identifies which values from a tool response
are reflected in the reference answer; the subsequent corruption prompts use
this information to create synthetic hallucinations. These prompts are not
used to annotate the naturally generated responses described above.

\begin{tcolorbox}[
    colback=gray!5!white,
    colframe=gray!75!black,
    title=Prompt for LLM-Judge,
    fonttitle=\bfseries,
    sharp corners,
    before skip=1pt,
    after skip=1pt
]

You are a careful evaluator analyzing how a tool response is reflected in an
assistant's answer. Your goal is to identify which fields from the tool
response were USED (mentioned/surfaced in the answer) versus UNUSED.

You will also judge whether the answer is appropriately selective or whether
it omits important information a complete answer should have included.

Definitions:

- USED field: the field's value (or a paraphrase/reformatting of it) appears
  in the answer, OR the answer directly answers a question with that value.
  
- UNUSED field: the field exists in the tool response but its content is
  neither mentioned nor reflected in the answer.
  
- UNDERGENERATING: the answer skips information that the user's question
  would naturally expect. For example, if the user asked for "details" and
  the answer mentions only 1 of 5 returned items, that's undergeneration.
  But if the user asked "what's the temperature" and the answer gives just
  the number while skipping metadata like coordinates and timestamps, that's
  appropriately selective, NOT undergeneration.

Output STRICT JSON with this exact shape:
\{
  "used\_fields": ["path.to.field1", "path.to.field2"],
  "unused\_fields": ["path.to.field3"],
  "is\_undergenerating": false,
  "reasoning": "Brief explanation (1-2 sentences)"
\}

Field paths use dot notation for nested objects (e.g., "results.data.price")
and bracket notation for array indices (e.g., "events[0].name"). For arrays
of homogeneous items, you may use a wildcard like "events[*].name" if every
item is treated the same way in the answer.

Respond with ONLY the JSON object, no markdown fences or other text.

\end{tcolorbox}

\begin{tcolorbox}[
    colback=gray!5!white,
    colframe=gray!75!black,
    title=Hallucination prompt,
    fonttitle=\bfseries,
    sharp corners,
    before skip=1pt,
    after skip=1pt
]
You are a data-generation assistant for an LLM evaluation benchmark.

Your task: given a tool's JSON response, produce a MODIFIED version of it
where some values have been changed so the data is factually different.

Rules:
1. Change at least 1 but no more than 3 distinct values (numbers, names,
   dates, statuses, enums, booleans).
   
2. Keep the JSON structure, keys, and schema IDENTICAL to the original.

3. Changed values must be the same type and plausible for the field.

4. Do NOT add or remove any keys or array items.

5. The changes should be subtle — not obviously absurd.

TARGETING (when target fields are specified):
If a list of TARGET FIELDS is provided, prefer to change values at those
field paths. These are the fields most likely to appear in the assistant's
answer, so changing them produces visible hallucinations. You may still
make changes outside the target list if it improves realism, but ALWAYS
modify at least one target field if any are given.

Respond with ONLY the modified JSON, no explanations or markdown fences.
\end{tcolorbox}

\begin{tcolorbox}[
    colback=gray!5!white,
    colframe=gray!75!black,
    title=Overgeneration prompt,
    fonttitle=\bfseries,
    sharp corners,
    before skip=1pt,
    after skip=1pt
]
You are a data-generation assistant for an LLM evaluation benchmark.

Your task: given a user question, a tool call, and the tool's JSON response,
generate an assistant reply to the LAST user question that correctly states
the tool results but also adds extra information NOT present in the tool
response.

If conversation history is provided, treat it as background only.

Rules:
1. First, faithfully convey the key data from the tool response.

2. Then add plausible-sounding but UNSUPPORTED details: historical context,
   general knowledge, tips, comparisons, or elaborations.
   
3. Do NOT contradict any values in the tool response.

4. Do NOT reference or offer to use any other tools or capabilities.

5. Keep the added content to 1-3 extra sentences.

6. Try to insert these details into different parts of the original sentence, not just at the end.

CRITICAL FORMATTING RULE:
Wrap ALL added/unsupported content in <OVER> tags. Content that faithfully
represents the tool response should NOT be wrapped.

Example:
  Tool response: \{"weather": "sunny", "temp": "25°C"\}
  Your output:  <OVER> This is typical for this region during summer.</OVER> The weather is sunny with a temperature of 25°C. <OVER>It is a \
great time to visit outdoor attractions.</OVER>

Respond with ONLY the annotated assistant message
\end{tcolorbox}

\begin{tcolorbox}[
    colback=gray!5!white,
    colframe=gray!75!black,
    title=Missing tool prompt,
    fonttitle=\bfseries,
    sharp corners,
    before skip=1pt,
    after skip=1pt
]

You are a data-generation assistant for an LLM evaluation benchmark.

Your task: given a user question, a tool call, and the tool's JSON response,
generate an assistant reply to the LAST user question that correctly states
the tool results but then offers to perform an action using a tool that
does NOT exist.

If conversation history is provided, treat it as background only.

CRITICAL: If an "ORIGINAL SYSTEM PROMPT" is provided, it lists the tools
that DO exist in this system. Your fabricated capability MUST be a tool
or service NOT listed there. Do NOT invent capabilities that match any
existing tool's purpose — pick something genuinely outside the available
toolset (booking, scheduling, sending messages, executing transactions,
generating media, etc., as long as no listed tool covers it).

Rules:

1. First, faithfully convey the key data from the tool response.

2. Then suggest an additional action requiring a DIFFERENT, non-existent
   tool or service. Make the offer sound natural and related to context.
   
3. Do NOT contradict any values from the tool response.

4. Do NOT add unsupported factual claims about the tool's data.

CRITICAL FORMATTING RULE:
Wrap ALL references to non-existent capabilities in <MISSING> tags.

Example:
  Available tools: Weather\_API
  Tool response: {"weather": "sunny"}
  Your output: The weather is sunny today. <MISSING>Would you like me to
book a restaurant with outdoor seating for you?</MISSING>

Respond with ONLY the annotated assistant message.
\end{tcolorbox}

\begin{tcolorbox}[
    colback=gray!5!white,
    colframe=gray!75!black,
    title=Undergeneration prompt,
    fonttitle=\bfseries,
    sharp corners,
    before skip=1pt,
    after skip=1pt
]
You are a data-generation assistant for an LLM evaluation benchmark.

Your task: given a user question, a tool call, and the tool's JSON response,
generate an assistant reply that omits significant RELEVANT information
from the tool response while appearing complete.

KEY DISTINCTION:
A good answer is naturally SELECTIVE — it skips metadata and irrelevant
fields. That is NOT undergeneration. Undergeneration means dropping
information the user actually wanted or that's central to answering their
question. Examples:

  Good selectivity (NOT undergeneration):
    User: "what's the temperature?" → answer skips coordinates, timestamp.
  Undergeneration:
    User: "list all events" → answer skips half the events.
    User: "weather forecast for tomorrow" → answer gives temperature
      but skips precipitation and wind which the forecast includes.
    User: "give me details on this product" → answer skips price.

Rules:

1. Identify what fields are RELEVANT to the user's question.

2. Drop at least one of those relevant fields or items, without
   acknowledging the omission.
   
3. If the tool returns a list of items the user asked to enumerate,
   drop at least one entire item.
   
4. Do NOT contradict any values you DO include.

5. Do NOT add information beyond what the tool returned.

6. Do NOT reference or offer to use any other tools.

7. The response should read as if it's complete — don't say "here are some"
   or hint that information was left out.

Additionally, output a JSON block at the END of your response listing
exactly what was omitted that should have been included, in this format:

<OMITTED>
{"omitted\_items": ["description of each relevant piece of information that was dropped"]}
</OMITTED>

Only list items that a complete answer to the user's question SHOULD
contain. Do not list metadata that would naturally be omitted.

Example:
  User: "Find me upcoming AI events"
  Tool response: \{"events": [{"id": 1, "name": "A", "date": "Jan 1"\}, \
\{"id": 2, "name": "B", "date": "Feb 2"\}]}
  Your output: Here is the upcoming AI event:

  1. **A** — January 1st

  <OMITTED>
  \{"omitted\_items": ["Event B (id: 2, date: Feb 2) was completely dropped — user asked for events plural"]\}
  </OMITTED>

Respond with ONLY the assistant message followed by the OMITTED block.

\end{tcolorbox}

\section{GPT-4o Evaluation Prompts}
\label{app:gpt-eval-prompts}

This section describes the prompts used to evaluate GPT-4o as a hallucination detector in the zero-shot and few-shot settings.

\subsection{System Prompt}
\label{app:gpt-system-prompt}

The following system prompt is shared by both the zero-shot and few-shot configurations.
In the few-shot setting, five in-context examples (one per hallucination type) are appended as user/assistant message pairs before the test instance.

\begin{tcolorbox}[
    colback=gray!5!white,
    colframe=gray!75!black,
    title=System prompt for hallucination detection,
    fonttitle=\bfseries,
    sharp corners,
    breakable,
    before skip=1pt,
    after skip=1pt,
]
You are an expert evaluator of tool-calling dialogues for hallucination detection.

You will be given a dialogue that includes:
- A system prompt describing available tools
- A conversation between a user, an assistant (which may call tools), and tool responses
- The final assistant answer to evaluate

Your task is to detect hallucinations in the FINAL assistant answer and classify the row.

\#\# Hallucination type definitions

- clean: The final answer is fully grounded in the tool output. No hallucinated content.

- answer\_mismatch: The answer directly CONTRADICTS or SUBSTITUTES values from the tool response. For example: the tool returned "Gonzaga" but the answer says "Miami"; the tool returned year 2022 but the answer says 2023. The tool response IS present -- the assistant misread or swapped its data.
SPAN RULE: mark ONLY the wrong value itself (e.g. "Miami", "2023", "7.5") -- NOT the words before it (not "the team is", not " is", not "SD: "). The span must be the minimal substituted token.

- missing\_tool: The answer contains fabricated concrete facts (IDs, names, amounts, dates) that are NOT present in or derivable from the tool response (Baseless Info); OR appends an unsolicited proactive offer / question not grounded in any tool result or user request (Proactive Follow-up, e.g. "I can help you with more actions if you'd like", "I'll go ahead and...", "Need a ... ?"). A tool response IS always present.
SPAN RULE for Proactive Follow-up: the hallucinated span is almost always the FINAL sentence(s) of the answer -- the unsolicited offer or question appended at the end. Look at the very end of final\_answer first. Do NOT mark content from the middle of the answer unless it contains fabricated facts (Baseless Info) that are absent from the tool response.

- overgeneration: The answer adds unsupported elaboration, commentary, interpretation, or claims that go beyond what the tool returned, without directly contradicting it.

- undergeneration: The answer OMITS information that the tool returned. It is an absence, not a fabrication. For this type, span\_labels MUST be empty [].

\#\# Output format

Respond with valid JSON and NOTHING else. Schema:

\{
\quad"type": "<>",
\quad"span\_labels": [
\quad\quad\{
\quad\quad\quad"start": <int, char offset in final\_answer>,
\quad\quad\quad"end": <int, char offset in final\_answer, exclusive>,
\quad\quad\quad"text": "<exact substring from final\_answer>"
\quad\quad\}
\quad],
\quad"reasoning": "<one or two sentence explanation>"
\}

Rules for span\_labels:
- Provide spans ONLY for types: answer\_mismatch, missing\_tool, overgeneration.
- For clean and undergeneration the list MUST be empty [].
- start and end are character offsets into final\_answer (0-indexed, end is exclusive).
- text must be the exact substring final\_answer[start:end].
- Mark only the MINIMAL hallucinated text -- do not over-span:
\quad* answer\_mismatch: mark only the wrong token/value, not surrounding words.
\quad* missing\_tool (Proactive Follow-up): mark the final unsolicited sentence(s) at the END.
- You may return multiple spans if there are multiple hallucinated regions.
\end{tcolorbox}

\subsection{User Message Template}
\label{app:gpt-user-template}

Each test instance is formatted as follows and passed as the user message:

\begin{tcolorbox}[
    colback=gray!5!white,
    colframe=gray!75!black,
    title=User message template,
    fonttitle=\bfseries,
    sharp corners,
    before skip=1pt,
    after skip=1pt,
]
Evaluate this dialogue for hallucinations.

System prompt: <system\_prompt>

Conversation:
[user]: <user\_message>
[assistant]: <assistant\_tool\_calls>
[tool]: <tool\_responses>
[assistant]: <final\_assistant\_answer>

final\_answer: "<final\_assistant\_answer>"
\end{tcolorbox}

\subsection{Few-Shot In-Context Examples}
\label{app:gpt-few-shot-examples}

In the few-shot setting, five user/assistant example pairs (one per hallucination type) are inserted between the system prompt and the test instance.
Table~\ref{tab:few-shot-examples} summarizes the examples used.

\begin{table}[h]
\centering
\small
\begin{tabular}{llp{6.5cm}}
\toprule
\textbf{\#} & \textbf{Type} & \textbf{Scenario} \\
\midrule
1 & \texttt{clean} & Web page to PDF conversion; all values grounded. \\
2 & \texttt{answer\_mismatch} & Final Four teams; year and team name substituted. \\
3 & \texttt{missing\_tool} & Token balances; unsolicited portfolio report offer appended. \\
4 & \texttt{overgeneration} & Projectile motion; editorial commentary about wind resistance added. \\
5 & \texttt{undergeneration} & Football events count; specific number ``15'' omitted from answer. \\
\bottomrule
\end{tabular}
\caption{Summary of few-shot in-context examples for GPT-4o evaluation.}
\label{tab:few-shot-examples}
\end{table}

\label{sec:appendix}

\end{document}